\documentclass[letterpaper, 10 pt, journal, twoside]{IEEEtran}
\usepackage{amsmath,amsfonts}
\usepackage{algorithmic}
\usepackage{algorithm}
\usepackage{array}
\usepackage[caption=false,font=normalsize,labelfont=sf,textfont=sf]{subfig}
\usepackage{textcomp}
\usepackage{stfloats}
\usepackage{url}
\usepackage{verbatim}
\usepackage{graphicx}
\usepackage{cite}
\usepackage{makecell}
\usepackage{color}
\usepackage{multirow}
\usepackage{amssymb}
\usepackage{booktabs}
\usepackage{colortbl}
\usepackage{pifont}
\definecolor{forestgreen}{RGB}{47, 159, 87}

\newcolumntype{g}{>{\columncolor[gray]{0.9}}c}
\begin{document}

\title{SEPT: Standard-Definition Map Enhanced Scene Perception and Topology Reasoning for Autonomous Driving}

\author{Muleilan Pei, Jiayao Shan, Peiliang Li, Jieqi Shi, Jing Huo, Yang Gao, and Shaojie Shen
\thanks{This work was supported in part by the Hong Kong PhD Fellowship Scheme, and in part by the HKUST-DJI Joint Innovation Laboratory. \textit{(Corresponding author: Jieqi Shi.)}}
\thanks{Muleilan Pei and Shaojie Shen are with the Department of Electronic and Computer Engineering, Hong Kong University of Science and Technology, Hong Kong, China (email: mpei@ust.hk; eeshaojie@ust.hk).}
\thanks{Jiayao Shan and Peiliang Li are with Zhuoyu Technology, Shenzhen, China (email: jiayao.shan@zyt.com; peiliang.li@zyt.com)}
\thanks{Jieqi Shi, Jing Huo, and Yang Gao are with State Key Laboratory for Novel Software Technology, Nanjing University, Nanjing, China (email: isjieqi@nju.edu.cn; huojing@nju.edu.cn; gaoy@nju.edu.cn).}
}
\markboth{IEEE Robotics and Automation Letters. Preprint Version. May, 2025}
{Pei \MakeLowercase{\textit{et al.}}: SEPT: Standard-Definition Map Enhanced Scene Perception and Topology Reasoning for Autonomous Driving}


\maketitle

\begin{abstract}
Online scene perception and topology reasoning are critical for autonomous vehicles to understand their driving environments, particularly for mapless driving systems that endeavor to reduce reliance on costly High-Definition (HD) maps. However, recent advances in online scene understanding still face limitations, especially in long-range or occluded scenarios, due to the inherent constraints of onboard sensors. To address this challenge, we propose a \underline{S}tandard-Definition (SD) Map \underline{E}nhanced scene \underline{P}erception and \underline{T}opology reasoning (SEPT) framework, which explores how to effectively incorporate the SD map as prior knowledge into existing perception and reasoning pipelines. Specifically, we introduce a novel hybrid feature fusion strategy that combines SD maps with Bird’s-Eye-View (BEV) features, considering both rasterized and vectorized representations, while mitigating potential misalignment between SD maps and BEV feature spaces. Additionally, we leverage the SD map characteristics to design an auxiliary intersection-aware keypoint detection task, which further enhances the overall scene understanding performance. Experimental results on the large-scale OpenLane-V2 dataset demonstrate that by effectively integrating SD map priors, our framework significantly improves both scene perception and topology reasoning, outperforming existing methods by a substantial margin. 
\end{abstract}


\section{Introduction}
\IEEEPARstart{S}{cene} understanding is essential for autonomous vehicles, facilitating critical downstream tasks such as accurate motion prediction and decision-making. High-Definition (HD) maps play a pivotal role in this process, providing rich geometric and semantic information, as well as topology relationships. However, HD maps present significant challenges, including high annotation costs, scalability limitations, and ongoing maintenance demands \cite{li2024local}, which underscore the increasing need for online scene perception and topology reasoning \cite{li2023graph}. 

In recent years, vision-centric mapless driving approaches (i.e., driving without HD maps) have made significant strides \cite{hu2023planning, jiang2023vad}, especially within advanced driver assistance systems. These methods aim to reduce the heavy reliance on HD maps by leveraging onboard sensors to perceive the complex scene structure of driving environments in real time. Specifically, with multi-view images as input, a variety of tasks need to be addressed, including lane segment detection, traffic element recognition, and scene topology reasoning \cite{wang2024openlane, li2023lanesegnet}. 

\begin{figure}[t]
    \centering
    \includegraphics[width=0.48\textwidth]{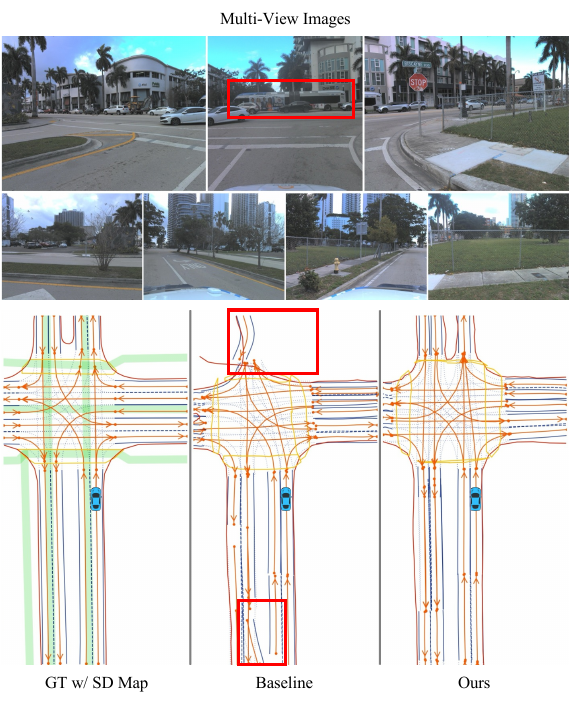}
    \vspace{-0.1cm}
    \caption{Illustration of the SD map prior for enhancing online scene understanding in long-distance and occlusion scenarios. In this example, the front view is severely obstructed by a bus at the intersection (highlighted in the red box in the front-view image), and the left-back zone is distant. The baseline (LaneSegNet \cite{li2023lanesegnet}) fails to correctly perceive the road structure (indicated by the red boxes in the top and bottom of the middle visualization), whereas our SEPT framework accurately predicts the road layout with the augmentation of the SD map prior. The Ground Truth (GT) of lane segments is shown in the left figure, with the green line representing the SD map.
}
    \label{Fig1}
    \vspace{-0.3cm}
\end{figure}

Nevertheless, due to the inherent limitations of onboard sensors, such as constrained perception range and restricted field of view, fully mapless driving systems often struggle to accurately reconstruct far-seeing or occluded road conditions. Given that human drivers typically perceive the surrounding scenarios by combining observations with navigation maps, also known as Standard-Definition (SD) maps \cite{ort2022maplite}, integrating SD maps as additional prior knowledge of road structures offers a promising solution to complement onboard sensory inputs. In general, the SD map provides a centerline skeleton of road networks without detailed and high-precision annotations \cite{jiang2024p}, making it lightweight, scalable, easily accessible, and low-cost \cite{zhang2024sdhdmap}. This basic geographic and road-level topological information can effectively augment online sensing capabilities, thereby enhancing scene perception and topology reasoning, particularly in long-distance or occlusion scenarios, as demonstrated in Fig. \ref{Fig1}.

Despite the substantial potential benefits of SD maps, the effective integration of such map priors into current online perception and reasoning paradigms remains an ongoing challenge. Existing approaches typically rely on relatively simple encoding strategies to represent SD maps, either in a rasterized \cite{jiang2024p} or vectorized \cite{luo2024augmenting} format. Each representation has distinct advantages: dense rasterization preserves spatial positional information and fine-grained local details, while sparse vectorization captures complex geometry and topology more efficiently. However, most methods either focus on one representation or combine the two in a simplistic manner \cite{yang2024toposd}, which limits effective feature extraction and results in suboptimal utilization or information loss from the SD map. To address this gap, we encode the SD map using a hybrid representation and propose a lightweight yet effective fusion module to augment the Bird’s-Eye-View (BEV) features with SD map priors.
Additionally, inherent inaccuracies in GPS signals often cause weak spatial misalignment between the SD map and BEV space \cite{wu2024maplocnet}. While previous works tend to neglect this artifact or dismiss it as noise, we introduce a feature alignment mechanism to resolve this issue. Specifically, for rasterization, we design a feature transformation network that dynamically modulates the features through predicting a scaling factor and bias term for each feature channel; for vectorization, we adopt a cross-attention mechanism \cite{vaswani2017attention} that adaptively attends to corresponding features, ensuring better alignment with the BEV feature space. 

Moreover, existing approaches overlook the importance of topological road structures in driving scenes. For example, intersections, including cross, merge, or diverge nodes, serve as critical topological attributes that signify changes in road networks. Such keypoints can be effectively identified from SD map priors, which provide valuable characteristics about road structures. To leverage this information, we introduce an auxiliary task focused on recognizing the distribution of road intersections derived from SD maps. This task enables BEV features to capture crucial road topology, thereby enhancing overall driving scene understanding.

In summary, the primary contributions of this letter are as follows: 
(1) We propose a novel hybrid fusion strategy for SD maps that combines both rasterized and vectorized representations, ensuring effective alignment with BEV features for improved synergy. 
(2) We introduce an auxiliary Intersection-aware KeyPoint Detection (IKPD) task conditioned on the SD map prior, further enhancing scene understanding capabilities.
(3) Extensive experiments on the large-scale OpenLane-V2 dataset demonstrate that our SD map-enhanced framework, termed SEPT, significantly improves both scene perception and topology reasoning performance.

\section{Related Work}
\subsection{Online Scene Perception}
Online HD map construction relies on the accurate perception of scene elements. Pioneering efforts have focused on laneline detection \cite{liao2022maptr, li2022hdmapnet} to capture road geometry, or centerline perception \cite{li2023graph, xu2023centerlinedet} to recognize lane connectivity. Given the intertwined nature of these two representations, a comprehensive mapping format, lane segment \cite{li2023lanesegnet}, has been proposed to seamlessly integrate both geometric 3D lanelines and topological 3D lane centerlines, along with areas defined by road boundaries and pedestrian crossings. Additionally, traffic element recognition has also been extensively explored in the literature \cite{langenberg2019deep, liu2022petr} for driving scene understanding, including the detection of traffic lights, road signs, and their associated semantic attributes. Despite advances in detecting these map elements, current online scene perception systems still struggle with occlusions and long-range scenarios. To address these limitations, our work leverages SD map priors, serving as essential complementary prompts with the potential to improve performance in these challenging conditions. 

\subsection{Scene Topology Reasoning}
Scene topology information is significant for downstream trajectory prediction \cite{liang2020learning} and behavior planning \cite{liureasoning} tasks, as it provides the topological relationships among lanes and between lanes and traffic elements. Nevertheless, research on topology reasoning has been limited until the emergence of the OpenLane-V2 benchmark \cite{wang2024openlane}, which utilizes adjacency matrices to characterize topological connectivity. Most existing methods rely on Multi-Layer Perceptrons (MLPs) \cite{can2021structured} or Graph Neural Networks (GNNs) \cite{li2023graph} to learn these connection relationships, or incorporate spatial position encoding \cite{wutopomlp} to enhance reasoning capabilities. These methods, however, are prone to disruption by endpoint shift issues. To address this, the calculation of geometric distance and semantic similarity \cite{fu2024topologic} has been proposed to mitigate such effects. Moreover, since SD maps inherently contain the topological structure of driving scenes, recent works \cite{ma2025roadpainter} have explored leveraging this prior knowledge to further improve topology reasoning.

\begin{figure*}[t]
    \centering
    \includegraphics[width=0.98\textwidth]{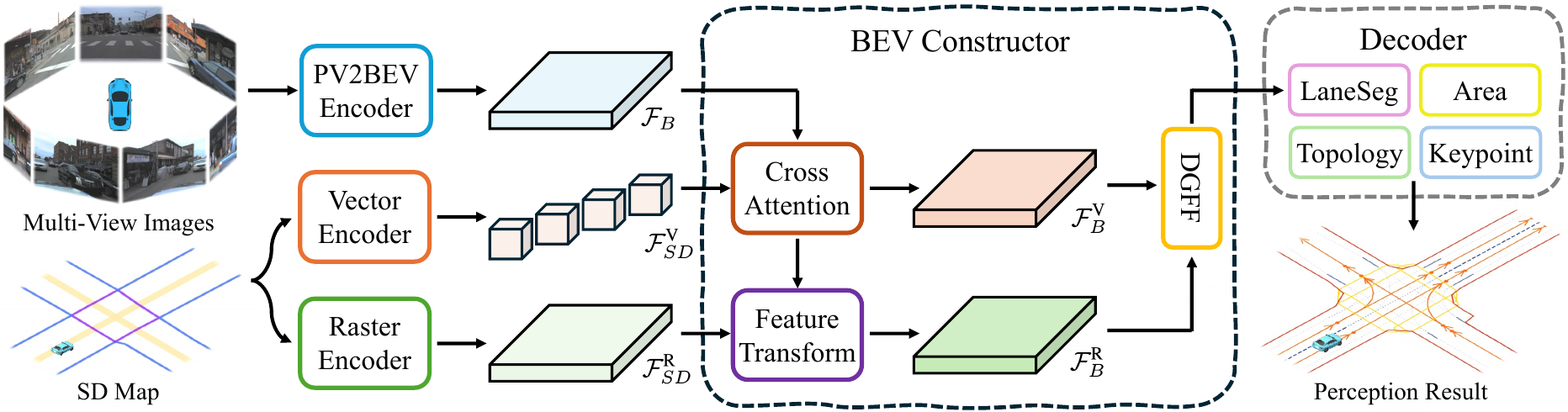}
    \caption{Overview of the SEPT architecture, demonstrating how it enhances the existing perception and reasoning model for online scene understanding through the integration of the SD map prior.
}
    \label{Fig2}
    \vspace{-0.4cm}
\end{figure*}

\subsection{SD Map Prior for Autonomous Driving}
SD maps, such as Google Maps, are widely used for urban navigation and have recently garnered increasing attention in autonomous driving tasks. Previous studies have primarily concentrated on leveraging SD map priors to enhance online map construction, particularly in long-range scenarios \cite{wu2024blos}. These methods typically involve rasterizing SD maps \cite{ort2022maplite} and employing Convolutional Neural Networks (CNNs) to extract features. However, the intrinsic weak alignment between SD maps and BEV features remains a challenge \cite{li2024local}, leading to the adoption of attention mechanisms \cite{jiang2024p}. Recent advances in topology reasoning have also incorporated SD maps by vectorizing them into polylines and using Transformer  \cite{luo2024augmenting} or GNN \cite{zhang2024sdhdmap} architectures to improve online lane topology understanding. To fully exploit both representations, a concurrent approach \cite{yang2024toposd} combines these two distinct streams; however, its fusion strategy remains overly simplistic, limiting effectiveness. Considering the existing constraints in SD map utilization, our work further explores their potential by developing a powerful hybrid fusion module and introducing an auxiliary intersection-aware keypoint forecasting task.

\section{Methodology}
\subsection{Task Statement}
The online driving scene understanding task involves both scene perception and topology reasoning, using multi-view images and the corresponding SD map priors as inputs. Scene perception includes detecting lane segments, drivable areas, and traffic elements. To be specific, lane segments comprise directed lane centerlines, left and right lane boundaries, and their associated line types (e.g., non-visible, solid, dashed). Drivable areas are represented by undirected curves or closed polygons corresponding to road boundaries and pedestrian crossings. Traffic elements encompass traffic lights and road signs visible in the front view, together with their relevant attributes.
For topology reasoning, the goal is to infer the topological relationships among lane segments and between lane segments and traffic elements. This topological information is typically modeled as a lane graph, where nodes represent lane segments or traffic elements, and edges signify connectivity relationships. An adjacency matrix is employed to characterize the lane graph.

\subsection{Framework Overview}
The overall pipeline of our SEPT framework is illustrated in Fig. \ref{Fig2}, which improves the baseline model by incorporating SD map priors. Specifically, given multi-view images, the PV2BEV encoder first extracts visual information via the image backbone and then transforms the Perspective-View (PV) features into the BEV feature, denoted as $\mathcal{F}_{B}$, by view transformation.
Additionally, the SD map prior is encoded in two distinct formats: rasterized features $\mathcal{F}_{SD}^{\text{R}}$ and vectorized features $\mathcal{F}_{SD}^{\text{V}}$, through a hybrid SD map encoding approach. These two representations are then leveraged to augment the BEV feature through a Feature Transformation (FT) module and a cross-attention network, respectively, producing the enhanced BEV features $\mathcal{F}_{B}^{\text{R}}$ and $\mathcal{F}_{B}^{\text{V}}$.
A lightweight yet effective Dual Gated Feature Fusion (DGFF) module is employed to fuse these two augmented features, generating the final enhanced BEV feature $\mathcal{F}_{B}^{\text{SD}}$. This feature is consequently decoded to address various subtasks by different heads, such as the lane segment head, area head, topology head, etc. Notably, we also introduce an additional keypoint head for an auxiliary task, which detects road intersections from SD maps, further enhancing scene understanding capabilities.

\subsection{Hybrid SD Map Encoding and Fusion}
To fully leverage SD map priors, we introduce a hybrid encoding approach, utilizing both rasterized and vectorized formats. These two representations are incorporated to enhance the BEV feature while ensuring implicit alignment between them. In addition, we design an efficient and effective fusion strategy to seamlessly integrate these features, thereby improving overall performance. 
Herein, let the BEV feature be represented as \( \mathcal{F}_{B} \in \mathbb{R}^{H \times W \times C} \), where \( H \) and \( W \) correspond to the spatial dimensions of the BEV perception range, and \( C \) denotes the feature dimension.

\subsubsection{Vectorized SD Map Encoding}
Given raw polylines of SD maps, we begin by uniformly resampling these sequences to obtain \( M \) segments. For each segment, we further evenly sample a fixed number of points. Following the structure of the classical vectorized method, SMERF \cite{luo2024augmenting}, we then vectorize the SD map and extract the initial vectorized feature \( \mathcal{F}_{SD}^{\text{V}} \in \mathbb{R}^{M \times C} \) using a Transformer-based encoder model. In this paradigm, spatial misalignment between the SD map tokens and the BEV space can be mitigated through a multi-head cross-attention mechanism. Here, the BEV feature acts as query tokens, while the SD map tokens serve as keys and values. This enables the BEV queries to adaptively aggregate relevant SD map tokens conditioned on a learnable attention distribution. As a result, we obtain implicitly aligned BEV features \( \mathcal{F}_{B}^{\text{V}} \in \mathbb{R}^{H \times W \times C} \), complemented by the vectorized SD map priors.

\subsubsection{Rasterized SD Map Encoding}
We first rasterize the SD map into an \( H \times W \) canvas with a binary representation, where each grid cell is assigned a value of 1 if occupied by a polyline, and 0 otherwise. Different road types, such as crosswalks and sidewalks, are encoded as separate channels. The original SD map features are then extracted using CNNs, yielding the rasterized feature \( \mathcal{F}_{SD}^{\text{R}} \in \mathbb{R}^{H \times W \times C} \).
Note that this feature may be weakly misaligned with the BEV space. To address this, motivated by the T-Net in PointNet \cite{qi2017pointnet}, we introduce a Feature Transformation (FT) module to align \( \mathcal{F}_{SD}^{\text{R}} \) with \( \mathcal{F}_{B}^{\text{V}} \) at the feature level. Specifically, we first project both features along the channel dimension and compute their feature difference \( \mathcal{F}_\Delta \in \mathbb{R}^{H \times W \times C} \), which represents a form of calibration error. We then apply a max-pooling operation on \( \mathcal{F}_\Delta \) to obtain the global context vector \( \mathcal{F}_\Delta^{\text{Global}} \in \mathbb{R}^{C} \).
Next, we leverage Feature-wise Linear Modulation (FiLM) \cite{perez2018film} to predict the scaling factor \( \gamma \in \mathbb{R}^{C} \) and the bias term \( \beta \in \mathbb{R}^{C} \) for each feature channel. Finally, these transformation parameters are applied to the rasterized feature \( \mathcal{F}_{SD}^{\text{R}} \), resulting in the enhanced BEV feature \( \mathcal{F}_{B}^{\text{R}} \in \mathbb{R}^{H \times W \times C} \), with implicit spatial alignment, as follows:
\begin{equation}
    \mathcal{F}_{B}^{\text{R}} = \gamma \odot \mathcal{F}_{SD}^{\text{R}} + \beta,
\end{equation}
where \( \odot \) denotes the Hadamard (element-wise) product, and all operations follow the broadcasting mechanism.

\begin{figure}[t]
    \centering
    \includegraphics[width=0.48\textwidth]{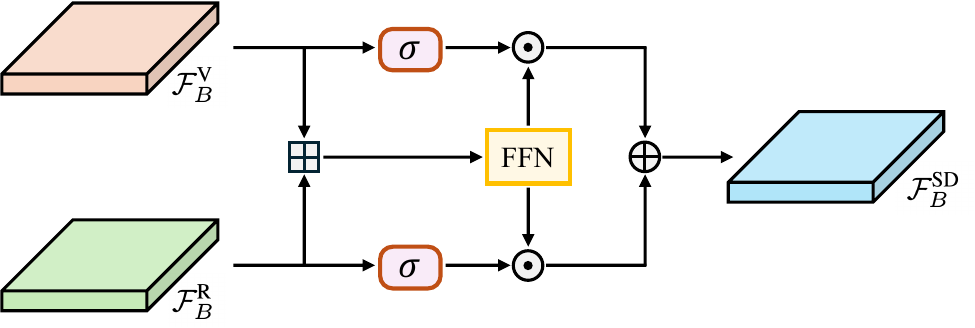}
    \caption{The hybrid feature fusion process of the DGFF module.}
    \label{Fig3}
    \vspace{-0.3cm}
\end{figure}

\subsubsection{Dual Gated Feature Fusion}
After obtaining the two BEV features augmented by rasterized and vectorized SD map features, it is essential to design an effective fusion strategy to combine these distinct features, as the characteristics of the two branches may differ significantly. To this end, we propose a lightweight yet powerful fusion network called the Dual Gated Feature Fusion (DGFF) module, which leverages the gated attention mechanism to aggregate the dual-branch features. As depicted in Fig. \ref{Fig3}, the two features \( \mathcal{F}_{B}^{\text{R}} \) and \( \mathcal{F}_{B}^{\text{V}} \) are first concatenated along the feature dimension and passed through a feed-forward network to produce a fused feature \( \mathcal{F}_{B}^{\text{R} \boxplus \text{V}} \in \mathbb{R}^{H \times W \times C} \), as follows:
\begin{equation}
     \mathcal{F}_{B}^{\text{R} \boxplus \text{V}} = \mathrm{FFN}(\mathcal{F}_{B}^{\text{R}} \boxplus \mathcal{F}_{B}^{\text{V}}),
\end{equation}
where \( \boxplus \) denotes concatenation along the feature dimension, and \( \mathrm{FFN}(\cdot) \) represents the feed-forward network.
Next, an element-wise gating mechanism is performed on each input stream using the sigmoid function, enabling the model to adaptively weight the contributions of rasterized and vectorized features. This is expressed as:
\begin{equation}
w_{\text{R}} = \sigma(\mathcal{F}_{B}^{\text{R}}), 
\quad
w_{\text{V}} = \sigma(\mathcal{F}_{B}^{\text{V}}),
\end{equation}
where \( \sigma(\cdot) \) is the element-wise sigmoid function, producing attention weights for each input stream. Although simpler, the gating mechanism introduces nonlinearity and adaptability, allowing the model to capture richer feature interactions without increasing the number of learnable parameters. This strikes a balance between representational capacity and efficiency.
Finally, two parallel projection networks refine each gated feature before merging them via a weighted averaging operation, generating the final enhanced BEV feature \( \mathcal{F}_{B}^{\text{SD}} \in \mathbb{R}^{H \times W \times C} \). This process can be formulated by the following expression:
\begin{equation}
\label{eq}
    \mathcal{F}_{B}^{\text{SD}} = \mu \cdot
    \mathrm{Proj}_\text{R} \bigl( w_{\text{R}} \odot \mathcal{F}_{B}^{\text{R} \boxplus \text{V}} \bigr) + \nu \cdot \mathrm{Proj}_\text{V} \bigl( w_{\text{V}} \odot \mathcal{F}_{B}^{\text{R} \boxplus \text{V}} \bigr),
\end{equation}
where \( \mathrm{Proj}_{\text{R}} \) and \( \mathrm{Proj}_{\text{V}} \) are the parallel projection networks. $\mu$ and $\nu$ are hyperparameters for balancing each term.

Overall, with the support of the DGFF module, our hybrid SD map encoding and fusion strategy can adaptively fuse heterogeneous feature representations, empowering the model to emphasize the most informative components from each branch. This substantially boosts the representation capabilities of both rasterized and vectorized SD map priors, while maintaining efficiency and expressiveness.

\subsection{Intersection-Aware Keypoint Detection}
To further enhance the BEV feature representation and improve understanding of road topology and geometry, we introduce an Intersection-Aware Keypoint Detection (IKPD) task. This auxiliary task helps the model capture road interaction patterns by detecting the road interaction distribution.

\subsubsection{Intersection Generation}
The first step in implementing the IKPD task involves identifying the intersection points of roads from SD maps, which will serve as the ground truth for supervision. Since the SD map prior provides the essential skeleton of road networks, intersection locations (e.g., merging, diverging, and crossing points) can be easily extracted.
However, intersection points are typically sparsely distributed across the scene, and directly using these points as supervision can lead to class imbalance during training. Additionally, due to intrinsic positional biases relative to the ground-truth HD maps in certain scenarios, the intersection points derived from the SD map may not perfectly align with the finer details of the road structure.
To mitigate this issue, we represent the keypoints as Gaussian distributions, similar to the approach used in confidence-based keypoint detection \cite{law2018cornernet, duan2019centernet}. Specifically, for each scene, we construct a heatmap \( \mathcal{H} \in \mathbb{R}^{H \times W \times 1} \) to model the ground-truth distribution of road intersections. Each intersection is represented as a Gaussian distribution centered at its location, with a certain radius reflecting the spatial uncertainty.

\subsubsection{IKPD Head}
Given the augmented BEV feature \( \mathcal{F}_{B}^{\text{SD}} \), we aim to design a lightweight network capable of effectively decoding the road intersection heatmap, thereby enriching the BEV feature with crucial geometric and topological information about the road structure. The IKPD head follows an efficient design paradigm that emphasizes both local feature extraction and global context reasoning.
Specifically, the BEV feature is first passed through a series of CNN blocks with residual connections. Each residual block comprises depthwise separable convolutions (i.e., a depthwise convolution followed by a pointwise convolution) \cite{chollet2017xception}, which decouple spatial and channel-wise operations for computational efficiency. Dilated convolutions are also incorporated for capturing broader spatial context information. After each convolution, the output feature is refined with the Squeeze-and-Excitation (SE) \cite{hu2018squeeze} attention, which recalibrates the channel-wise features by computing global statistics and adaptively weighting the importance of each channel. This allows the IKPD head to prioritize the most relevant features for keypoint detection, improving its ability to focus on critical patterns. Consequently, the final output is produced through a \( 1 \times 1 \) convolution followed by a sigmoid activation function, generating a heatmap that represents the distribution of road intersections.

\begin{table*}[t]
    \caption{Quantitative results on the OLV2 validation split, benchmarked using OLS. All metrics follow the higher-the-better criterion. The official ranking metric is shaded in gray, and the best results are indicated in \textbf{bold}. A "-" denotes the absence of relevant data. }
    \centering
    \setlength{\tabcolsep}{1.5mm}{
    \begin{tabular}{l|cc|ccg|ccg|c}
    \toprule
    \multirow{2}{*}{Method}  & \multirow{2}{*}{DET$_{l}\uparrow$} & \multirow{2}{*}{DET$_{t}\uparrow$} & \multicolumn{3}{c|}{v1.0} & \multicolumn{3}{c|}{v1.1} & \multirow{2}{*}{Params}\\ 

     & & & TOP$_{ll}\uparrow$ & TOP$_{lt}\uparrow$ & OLS$\uparrow$ & TOP$_{ll}\uparrow$ & TOP$_{lt}\uparrow$ & OLS$\uparrow$ \\
    \midrule
    
    TopoNet \cite{li2023graph}            & 28.6  & 48.6  & 4.1 & 20.3 & 35.6  & 10.9     & 23.8     & 39.8  & 62.6M \\
    w/ OLV2 \cite{zhang2024sdhdmap}    &  27.9 & 48.1 & 5.1 & 20.9 & 36.1 & - & - & - & 75.9M \\
    w/ OSMG \cite{zhang2024sdhdmap}    &  30.0 & 47.6 & 5.4 & 21.3 & 36.7 & - & - & - & 64.6M \\
    w/ OSMR \cite{zhang2024sdhdmap}  & 30.6  & 44.6 & 7.7 & 22.9 & 37.7 & - & - & - & 75.9M \\
    w/ SMERF \cite{luo2024augmenting}      & 33.4  & 48.6 & 7.5&  23.4 & 39.4 & 15.4  & 25.4 & 42.9 & 65.8M \\
    w/ \textbf{SEPT (Ours)}   & \textbf{34.2 (+5.6)}  & \textbf{49.8 (+1.2)} & \textbf{8.3 (+4.2)} & \textbf{23.8 (+3.5)} & \textbf{40.4 (+4.8)} & \textbf{19.5 (+8.6)}   & \textbf{27.5 (+3.7)}  & \textbf{45.2 (+5.4)} & 70.4M \\
    \midrule
    TopoLogic \cite{fu2024topologic}   & 29.2  & 46.5  & 18.0 & 20.6 & 40.9  & 23.6     & 24.2    & 43.4 & 61.8M \\
    w/ SMERF \cite{luo2024augmenting}  & 31.0  & 48.7  & 21.2 & 22.4 & 43.3  & 26.9     & 26.2     & 45.7 & 65.1M \\
    w/ \textbf{SEPT (Ours)}   & \textbf{34.3 (+5.1)}  & \textbf{48.9 (+2.4)} & \textbf{25.1 (+7.1)} & \textbf{25.1 (+4.5)} & \textbf{45.8 (+4.9)} & \textbf{31.2 (+7.6)}   & \textbf{29.7 (+5.5)}  & \textbf{48.4 (+5.0)} & 69.6M \\
    \bottomrule
  \end{tabular}
  }
  \label{table1}
  \vspace{-0.2cm}
\end{table*}

\subsection{Training Objectives}
Following the baseline approaches \cite{li2023graph, li2023lanesegnet}, the supervision is applied to each head to optimize distinct training objectives, including detection losses for lane segments, areas, and traffic elements, denoted as $\mathcal{L}_\text{DET}$, and topology reasoning losses, denoted as $\mathcal{L}_\text{TOP}$. 
Our proposed SEPT framework does not modify the baseline loss functions but introduces an additional loss term for the auxiliary IKPD head, denoted as $\mathcal{L}_\text{IKPD}$. Given the road intersection distribution is sparse and imbalanced, we employ focal loss \cite{law2018cornernet} to supervise the keypoint heatmap training. The overall loss $\mathcal{L}$ for SEPT is formulated as:
\begin{equation}
\mathcal{L} = \mathcal{L}_\text{DET} + \mathcal{L}_\text{TOP} +\mathcal{L}_\text{IKPD}.
\end{equation}

\section{Experiments and Results }
\subsection{Experiment Setups}
\subsubsection{Real-World Dataset}
We train and evaluate our proposed approach on the large-scale OpenLane-V2 (OLV2) dataset \cite{wang2024openlane}, which, to the best of our knowledge, is currently the only benchmark for both scene perception and topology reasoning in autonomous driving. All experiments in our work are conducted on the primary subset of OLV2, \textit{subset\_A}, built upon the Argoverse 2 \cite{wilson2023argoverse2} dataset with additional annotations for lane segments, traffic elements, and lane topology, etc. The \textit{subset\_A} comprises 1k scenes collected from six cities, with 2Hz multi-view images and optional SD map information (including three categories: roads, crosswalks, and sidewalks) extracted from the OpenStreetMap (OSM) \cite{haklay2008openstreetmap}. The training set consists of approximately 27k frames, while the validation set contains around 4.8k frames.

\subsubsection{Evaluation Metrics}
We evaluate the performance of perception and reasoning using the official metrics provided by OLV2 \cite{wang2024openlane}. There are two primary benchmark categories, each with distinct evaluation metrics: OLV2 Score (OLS) and OLV2 UniScore (OLUS). Both scores are averages derived from various metrics across different subtasks. The main distinction between them is that OLS focuses exclusively on lane centerline perception, while OLUS emphasizes lane segment perception. 
Specifically, OLS includes four sub-metrics: DET$_{l}$, DET$_{t}$, TOP$_{ll}$, and TOP$_{lt}$. DET$_{l}$ measures the mean average precision (mAP) for lane centerline detection, based on the Fréchet distance with match thresholds of 1.0, 2.0, and 3.0. DET$_{t}$ represents the mAP for traffic element recognition, conditioned on the average Intersection over Union (IoU) with a match threshold of 0.75 across various traffic attributes. TOP$_{ll}$ and TOP$_{lt}$ measure mAP for topology among lane centerlines and between lane centerlines and traffic elements, using the adjacency matrix. Notably, there are two versions for calculating TOP scores: V1.0 and V1.1, with the V1.0 calculation containing a potential loophole issue \cite{wutopomlp}. The OLS score is computed as the average of these four metrics, given by:
\begin{equation}
\text{OLS} = \frac{1}{4}\left[\text{DET}_l + \text{DET}_t + f\left(\text{TOP}_{ll} \right) + f\left(\text{TOP}_{lt} \right) \right],
\end{equation}
where \( f \) represents the square root function.

In contrast, OLUS encompasses five sub-metrics: DET$_{ls}$, DET$_{a}$, DET$_{te}$, TOP$_{lsls}$, and TOP$_{lste}$, covering detection for lane segments, areas, and traffic elements, as well as topology reasoning among lane segments and between lane segments and traffic elements. These metrics follow a similar calculation procedure to OLS, with the addition of DET$_{a}$, which is measured using Chamfer distance.

\subsubsection{Implementation Details}
We select two representative high-performance models as baselines: TopoNet \cite{li2023graph} for OLS and LaneSegNet \cite{li2023lanesegnet} for OLUS. To ensure a fair comparison, we retain the official implementations of both baseline models, incorporating only the modules designed specifically for the SD map prior into the codebase. All models employ the default ResNet-50 backbone, with BEV feature dimensions set to \( H = 200 \) and \( W = 100 \). We train our model on eight GPUs with a total batch size of 8. The training configuration, including the learning rate and optimizer, remains consistent with baseline settings, and all models share the same hyperparameters. 

\begin{table*}[tbp]
    \centering
    \caption{Quantitative results on the OLV2 validation split with map element buckets, benchmarked using OLUS.}
    \label{comparison_v2.1}
    \setlength{\tabcolsep}{1.5mm}
    \begin{tabular}{l|ccc|ccccc g| c}
        \toprule
        Method  & Raster & Vector & IKPD & DET$_{ls}\uparrow$ & DET$_a\uparrow$ & DET$_{te}\uparrow$  & TOP$_{lsls}\uparrow$  & TOP$_{lste}\uparrow$ & OLUS $\uparrow$ & Params\\
        \midrule    
        LaneSegNet \cite{li2023lanesegnet}	 &&&   & 30.9 & 20.0 & 36.7 & 25.6 & 20.8 & 36.7 & 61.8M \\
        w/ Raster Only & \ding{51} && & 33.8 (+2.9) & 28.1 (+8.1) & 38.1 (+1.4) & 27.5 (+1.9) & 21.8 (+1.0) & 39.9 (+3.2) & 65.5M \\
        w/ Vector Only & & \ding{51}& & 35.3 (+4.4) & 22.3 (+2.5) & 39.2 (+2.3) & 30.2 (+4.6) & 22.6 (+1.8) & 39.9 (+3.2) & 65.8M \\
        w/ Hybrid Fusion &\ding{51}&\ding{51}& & 35.8 (+4.9) & 28.2 (+8.2) & 39.6 (+2.9) & 31.0 (+5.4) & 22.7 (+1.9) & 41.4 (+4.7) & 69.8M \\
        w/ \textbf{SEPT} &\ding{51}&\ding{51}& \ding{51} & \textbf{38.4 (+7.5)} & \textbf{29.0 (+9.0)} & \textbf{40.0 (+3.3)} & \textbf{32.2 (+6.6)} & \textbf{23.8 (+3.0)} &  \textbf{42.6 (+5.9)} & 70.4M
        \\
        \bottomrule
    \end{tabular}
\vspace{-0.2cm}
\end{table*}

\subsection{Comparison with State-of-the-Art}
We compare our SEPT framework with other state-of-the-art methods that incorporate SD maps as input on the OLV2 benchmark, using the OLS evaluation metric. TopoNet \cite{li2023graph} serves as the baseline for centerline perception. TopoNet with OLV2, OSMG, and OSMR \cite{zhang2024sdhdmap} represent approaches utilizing rasterized SD maps from OLV2, augmented with full OSM attributes (e.g., stop signs, speed limits), and graph-based SD maps with OSM augmentation, respectively. SMERF is a classical method that enhances lane-topology understanding with SD maps in a vectorized representation. We present results for both v1.0 and v1.1 metrics to provide a comprehensive comparison, as shown in the upper group of Tab. \ref{table1}.
Since the v1.1 metric is a recent update, the results for TopoNet and SMERF on v1.1 have been reevaluated using their official checkpoints, while others are not available.
Compared to the baseline, our method significantly improves performance across all subtasks by effectively integrating SD map priors without introducing excessive parameters. Specifically, we achieve a 4.8 OLS improvement for v1.0 and a 5.4 OLS improvement for v1.1, with a notable 8.6 TOP$_{ll}$ increase in topology reasoning, outperforming other existing methods augmented with SD maps.
Moreover, we also apply our SEPT framework to the recent model, TopoLogic \cite{fu2024topologic}, which is designed to enhance lane topology reasoning. As shown in the lower group of Tab. \ref{table1}, our SEPT consistently improves performance across all subtasks, achieving a 4.9 OLS gain for v1.0 and a 5.0 OLS gain for v1.1.

In addition, we further evaluate our approach using the latest and more challenging OLUS evaluation metric, which focuses on lane segment perception and provides a more comprehensive assessment of the map element bucket. We directly apply our SEPT framework to LaneSegNet \cite{li2023lanesegnet}, a leading method for driving scene topology. As shown in Tab. \ref{comparison_v2.1}, even without further adaptation, our framework significantly enhances the baseline performance across all five subtasks, resulting in an overall improvement of 5.9 OLUS. This underscores the effectiveness and generalizability of our proposed framework.

\begin{table}[tbp]
    \centering
    \caption{Effect of FT module on rasterized SD map encoding.}
    \label{alignment}
    \setlength{\tabcolsep}{1.2mm}
    \begin{tabular}{c|ccccc g}
        \toprule
        FT   & DET$_{ls}$ & DET$_a$ & DET$_{te}$  & TOP$_{lsls}$  & TOP$_{lste}$ &  OLUS  \\
        \midrule
        \ding{55}   & 32.2 & 24.5 & 36.7 & 26.7 & 21.2  & 38.2 \\
        \ding{51}   & \textbf{33.8} & \textbf{28.1} & \textbf{38.1} & \textbf{27.5} & \textbf{21.8} & \textbf{39.9}  \\
        \bottomrule
    \end{tabular}
    \vspace{-0.2cm}
\end{table}

\subsection{Ablation Study} 
We conduct ablation studies to validate the effectiveness of each proposed component of SEPT as well as FT and DGFF modules using the OLV2 validation split, employing the OLUS evaluation metric for a comprehensive assessment. LaneSegNet \cite{li2023lanesegnet} serves as the baseline model.
\subsubsection{Component Study of SEPT} \label{CS}
We conduct an in-depth analysis of the contributions of each proposed component, as shown in Tab. \ref{comparison_v2.1}.

\noindent \textbf{Hybrid Fusion.} Compared to the baseline, integrating the SD map prior, whether in rasterized or vectorized representation, improves performance across all metrics, with both formats yielding comparable results. Specifically, the rasterized SD map significantly enhances area detection by 8.1 DET$_a$, as rasterization captures more spatial information. In contrast, the vectorized format improves the lane segment detection and topology reasoning between lane segments by 4.4 DET$_{ls}$ and 4.6 TOP$_{lsls}$, respectively, as vectorization better preserves the geometry and topology of the road structure. This highlights that both representations have distinct advantages and should complement each other, underscoring the superiority of employing our hybrid fusion strategy.

\noindent \textbf{IKPD.} As shown in the last two rows of Tab. \ref{comparison_v2.1}, incorporating the auxiliary IKPD task enables our method to fully exploit the potential of the SD map prior, leading to further improvements across all subtasks in scene perception and topology reasoning.

\begin{table}[tbp]
    \centering
    \caption{Comparison of different feature fusion strategies.}
    \label{fusion strategy}
    \setlength{\tabcolsep}{1.2mm}
    \begin{tabular}{l|ccccc g}
        \toprule
        Fusion Strategy  & DET$_{ls}$ & DET$_a$ & DET$_{te}$  & TOP$_{lsls}$  & TOP$_{lste}$ & OLUS \\
        \midrule
        Addition    & 35.4 & 23.9 & 36.8 & 29.5 & 21.7 & 39.4  \\
        Concatenation      & 35.7 & 24.3 & 37.7 & 29.9 & 22.0 & 39.9  \\
        Cross-Attention  & 35.5 & 25.5 & 38.3 & 30.0 & 22.5 & 40.3 \\
        DGFF (Ours) & \textbf{35.8} & \textbf{28.2} & \textbf{39.6} & \textbf{31.0} & \textbf{22.7} & \textbf{41.4} \\
        \bottomrule
    \end{tabular}
    \vspace{-0.1cm}
\end{table}

\begin{table}[tbp]
    \centering
    \caption{Comparison of different fusion weights.}
    \label{weight}
    \setlength{\tabcolsep}{1.2mm}
    \begin{tabular}{cc|ccccc g}
        \toprule
        Raster  & Vector   & DET$_{ls}$ & DET$_a$ & DET$_{te}$  & TOP$_{lsls}$  & TOP$_{lste}$ &  OLUS \\
        \midrule
        0.2 & 0.8	& 35.5 & 27.1 & 39.2 & 30.1 & 22.4 & 40.8 \\
        0.5 & 0.5   &\textbf{35.8} & \textbf{28.2} & \textbf{39.6} & \textbf{31.0} & \textbf{22.7} & \textbf{41.4} \\
        0.8 & 0.2  & 35.2 & 27.5 & 38.9 & 30.1 & 22.3 & 40.7 \\
        \bottomrule
    \end{tabular}
    \vspace{-0.2cm}
\end{table}

\subsubsection{Effect of the FT Module}  
We first investigate the impact of spatial alignment in the rasterized SD map encoding process. In the rasterized SD map-only configuration, we remove the proposed FT module from the pipeline. As shown in Tab. \ref{alignment}, the model with the FT module consistently outperforms its counterpart across all metrics. This highlights the importance of aligning the SD map and BEV feature space, with feature-level alignment effectively mitigating associated challenges.

\begin{figure*}[!t]
    \centering
    \includegraphics[width=\textwidth]{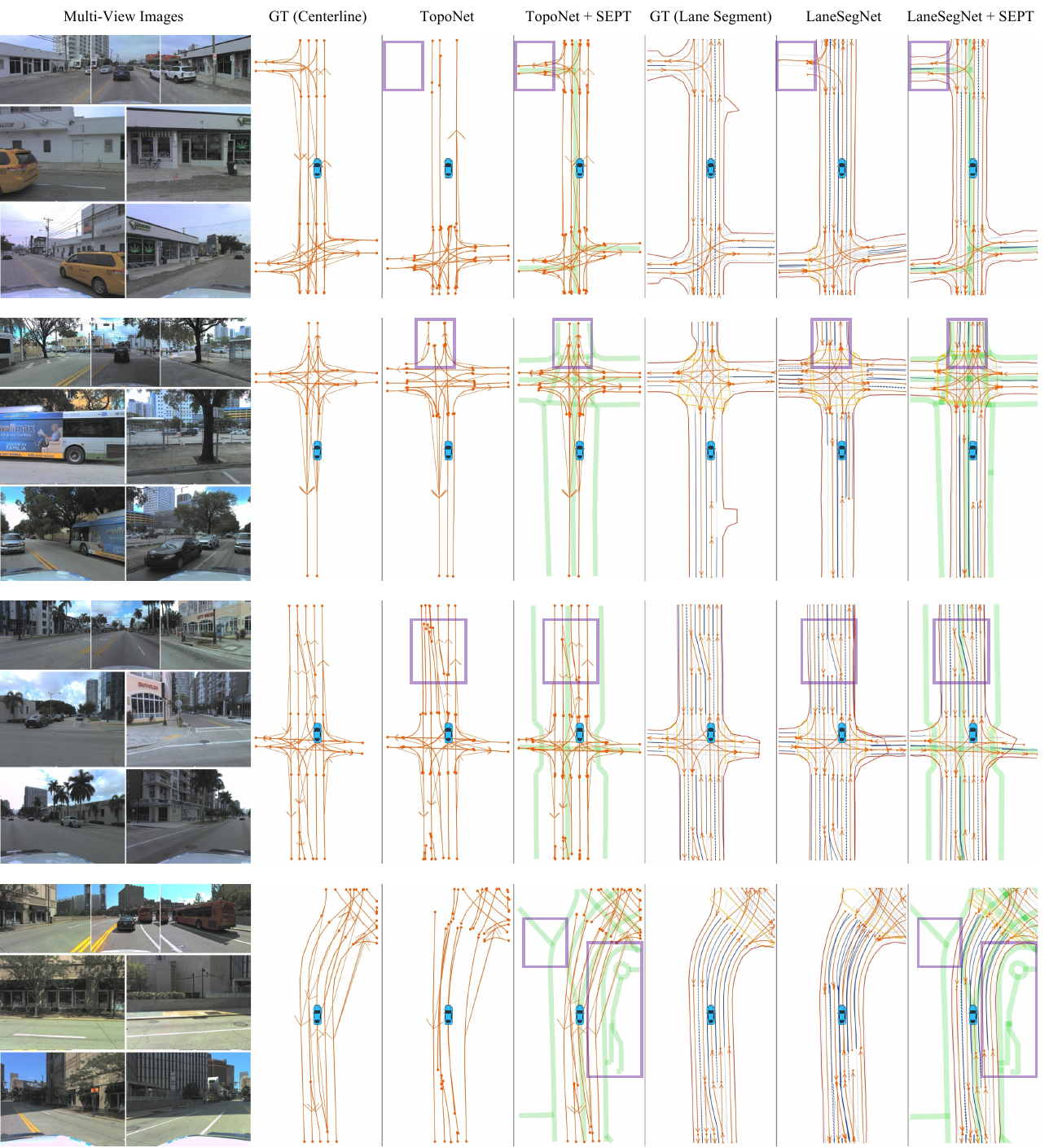}
    \caption{Qualitative comparisons between the baselines with and without our SEPT module on the OLV2 validation split. From left to right, the figure presents the multi-view images, the ground truth (GT) for centerline perception and topology, the results of the baseline (TopoNet) with and without SEPT, the GT for lane segment perception and topology, and the results of the baseline (LaneSegNet) with and without SEPT. The green line indicates the corresponding SD map prior.}
    \label{Fig4}
\vspace{-0.1cm}
\end{figure*}

\subsubsection{Effect of the DGFF Module}  
We further evaluate the fusion capability of the proposed DGFF module. Given the two BEV features \( \mathcal{F}_{B}^{\text{R}} \) and \( \mathcal{F}_{B}^{\text{V}} \), enhanced by rasterized and vectorized SD maps, respectively, we replace the DGFF with various fusion strategies, including element-wise addition, concatenation (followed by FFN), and cross-attention. As shown in Tab. \ref{fusion strategy}, simple combinations of these features either hinder overall performance or fail to achieve a synergistic effect. While cross-attention improves performance, it introduces significant computational overhead. In contrast, our DGFF module is both more efficient and effective, utilizing a gated attention mechanism to integrate the two features and deliver superior performance.
Additionally, we also explore the impact of different combination weights (i.e., $\mu$ and $\nu$ in Eq. \ref{eq}) for the gated rasterized and vectorized features. From Tab. \ref{weight}, we observe that a balanced combination of the two features yields the best performance, with the rasterized term slightly outperforming in area detection and the vectorized term excelling in lane segment detection. These results are consistent with the findings discussed in Section \ref{CS}.

\subsection{Qualitative Results}
We present visualizations from the OLV2 validation split to demonstrate the improvements brought by our proposed SEPT framework over two baseline models: TopoNet \cite{li2023graph} for lane centerline perception and LaneSegNet \cite{li2023lanesegnet} for lane segment perception. As illustrated in Fig. \ref{Fig4}, we highlight several representative cases. In the first row, we demonstrate that in long-range scenarios, the baseline models either fail to detect or inaccurately predict the left turn (highlighted in purple boxes), whereas our model successfully identifies it. In the second row, due to occlusion from a front vehicle, our approach notably enhances the prediction of the occluded road structure at the intersection. The third row illustrates how effectively incorporating the SD map improves lane topology reasoning. The final row presents an interesting case where the SD map provides outdated information, resulting in incorrect prior knowledge. However, our SEPT framework demonstrates the ability to prioritize online perception, yielding a correct prediction that aligns with current observations. This shows that our model strikes an effective balance between onboard sensing and SD map priors. Overall, our SEPT framework significantly improves both scene perception and topology reasoning. More qualitative results can be found in our supplementary video.

\section{Conclusion}
In this letter, we present SEPT, a novel framework that integrates SD map priors into existing perception and reasoning models to enhance online scene understanding. SEPT effectively mitigates the misalignment issue in both rasterized and vectorized SD map representations and leverages the DGFF module to fuse these features for synergistic improvement. To further exploit the potential of SD maps, we introduce the auxiliary IKPD task, which enhances the model in capturing road interaction patterns. We apply our SEPT to two baseline methods and validate it on the large-scale OLV2 benchmark. The significant performance gains demonstrate the superiority of our framework. Future work will focus on incorporating additional SD map prior information, such as lane numbers and road directions, to further enhance scene perception and topology reasoning for mapless driving.

\bibliographystyle{IEEEtran}
\bibliography{main}

\end{document}